%% file: root.tex
\documentclass[letterpaper, 10 pt, conference]{ieeeconf}
\usepackage{bm}
\usepackage{color,colortbl}
\usepackage{nicefrac}
\usepackage{caption}
\usepackage{csquotes}  
\usepackage{verse}

\usepackage{times}
\usepackage{epsfig}
\usepackage{graphicx}
\usepackage{amsmath}
\usepackage{amssymb}
\usepackage{gensymb}
\usepackage{subfig}
\usepackage{float}
\usepackage{mathtools}
\usepackage{dblfloatfix}

\usepackage{multirow}
\usepackage{array}
\usepackage{booktabs}
\usepackage[export]{adjustbox}
\usepackage{xfrac}
\usepackage{stmaryrd}

\usepackage{comment}
\usepackage{multirow}
\usepackage{array}
\usepackage{booktabs}
\usepackage{xspace}

\newcommand{\Acronym}[0]{SESC\xspace} 
\newcommand{\AcronymInt}[0]{SESC(\emph{I})\xspace} 

\makeatletter
\let\NAT@parse\undefined
\makeatother

\usepackage[pagebackref=true,breaklinks=true,letterpaper=true,colorlinks,bookmarks=false]{hyperref}

\begin{document}
%
\title{Robust Self-Supervised Extrinsic Self-Calibration}
%
%
%


\author{Takayuki Kanai, Igor Vasiljevic, Vitor Guizilini, Adrien Gaidon, and Rares Ambrus%
\\ Toyota Research Institute (TRI), Los Altos, CA
\\ {\tt \{first.lastname\}@tri.global}
}
\markboth{}
{} 

%



\input{figures/teaser_and_title.tex}


\begin{abstract}
\input{sections/abstract}
\end{abstract}

\section{Introduction}
\input{sections/introduction}

\section{Related Work}
\input{sections/related}

\section{Methodology}
\input{sections/methodology}

\section{Experiments}
\input{sections/experiments}

\section{Conclusion}

\input{sections/conclusion}
\bibliographystyle{IEEEtran}
\bibliography{references}


%








\end{document}

%% file: figures/teaser_and_title.tex
    \twocolumn[{
    \renewcommand\twocolumn[1][]{#1}
    \maketitle
    \vspace{-8mm}
    \begin{center}
        \centering
        \captionsetup{type=figure}
        \subfloat[Self-supervised learning with velocity supervision]{
            \includegraphics[width=0.33\textwidth]{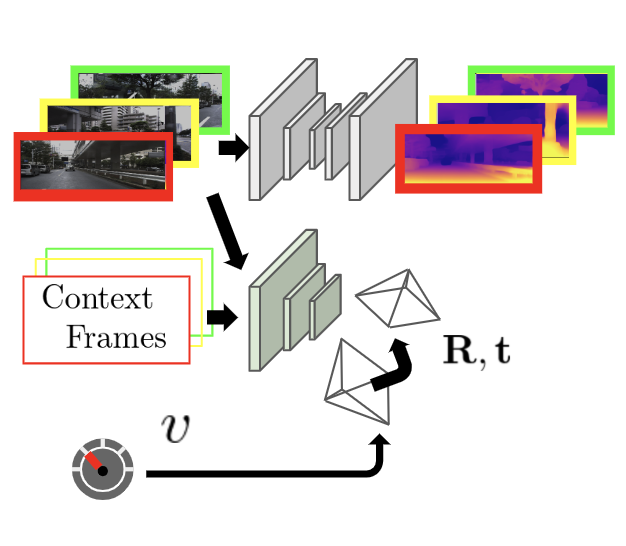}
            }
        \subfloat[Extrinsic estimation]{
            \includegraphics[width=0.30\textwidth]{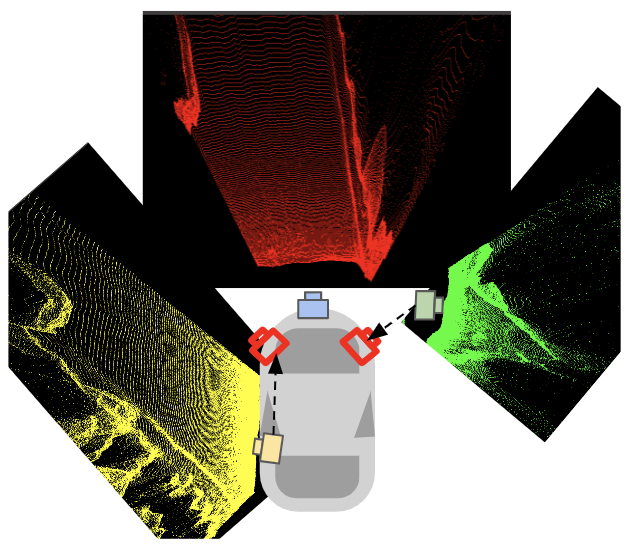}
            }
        \subfloat[Self-calibration via joint optimization]{
            \includegraphics[width=0.30\textwidth]{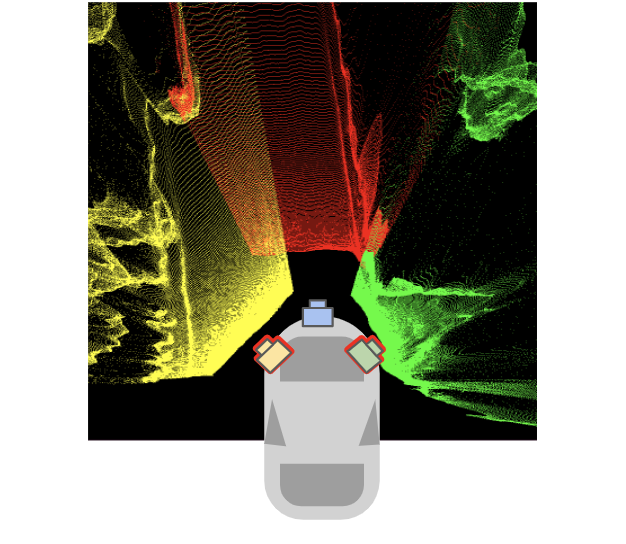}
            }
    \caption{
        \textbf{Our proposed method for self-supervised  extrinsic self-calibration (\Acronym).}
        (a) First, scale-aware depth and ego-motion predictors are acquired via self-supervised learning with instantaneous velocity supervision. (b) Afterwards, the extrinsic parameters are optimized using predictions from the learned depth and ego-motion networks. (c) Finally, all the networks and extrinsic parameters are jointly optimized to complete the self-calibration. 
    }
    \label{fig:teaser}
    \end{center}
    \vspace{1mm}
    }]

%% file: sections/abstract.tex
Autonomous vehicles and robots need to operate over a wide variety of scenarios in order to complete tasks efficiently and safely.
Multi-camera self-supervised monocular depth estimation from videos is a promising way to reason about the environment, as it generates metrically scaled geometric predictions from visual data without requiring additional sensors.
However, most works assume well-calibrated extrinsics to fully leverage this multi-camera setup, even though accurate and efficient calibration is still a challenging problem. 
In this work, we introduce a novel method for extrinsic calibration that builds upon the principles of self-supervised monocular depth and ego-motion learning.
Our proposed curriculum learning strategy uses monocular depth and pose estimators with velocity supervision to estimate extrinsics, and then jointly learns extrinsic calibration along with depth and pose for a set of overlapping cameras rigidly attached to a moving vehicle.
Experiments on a benchmark multi-camera dataset (DDAD) demonstrate that our method 
enables self-calibration in various scenes robustly and efficiently compared to a traditional vision-based pose estimation pipeline. 
Furthermore, we demonstrate the benefits of extrinsics self-calibration as a way to improve depth prediction via joint optimization. 
The project page: \href{https://sites.google.com/view/tri-sesc}{https://sites.google.com/view/tri-sesc}

%% file: sections/introduction.tex
Having access to accurate geometric information surrounding an autonomous system is crucial for perception and planning tasks.
Predicting dense depth maps from images is a promising way to generate this surrounding geometric information, as cameras are cheap, have low power consumption, and can be easily placed anywhere. Images are also useful for a number of other downstream tasks, such as detection, semantic segmentation, and tracking.
To achieve the goal of dense depth prediction, self-supervised depth and ego-motion learning methods~\cite{zhou2017unsupervised,monodepth2} are attractive because self-supervision is not restricted to high-quality labeled data, and can instead be trained on raw image sequences without ground-truth point clouds.
Moreover, to cover a wide field of view, research to extend the methods to a multi-camera setting has also emerged~\cite{fsm20222,kim2022selfsupervised} as this learning principle is not restricted to the number of cameras.
Importantly, studies have demonstrated that even if the overlap between two images from different cameras is small, provided that accurate extrinsics are available, multi-camera constraints still lead to improvements over monocular baselines~\cite{fsm20222}.

However, despite the fact that multi-camera training improves self-supervised learning, it still requires \textit{known} extrinsics, which in turn requires laborious camera calibration, typically involving the manual collection of images containing known calibration targets~\cite{Chen2022extrinsic}.
In contrast to the manual, target-based calibration approach, recent work has shown that monocular intrinsics calibration can be achieved in a self-supervised way~\cite{gordon2019depth,vasiljevic2020neural,tri-selfcalibration}, though self-supervised architectures utilizing extrinsics still use ground-truth data.
To tackle the problem, in this paper we propose \textbf{Self-Supervised Extrinsic Self-Calibration} (\Acronym): a novel methodology for \textit{extrinsic} self-calibration that extends the self-supervised depth and ego-motion learning paradigm. Utilizing pretrained scale-aware depth networks (obtained using the vehicle’s instantaneous velocity~\cite{packnet,fisheye2020}), and a curriculum learning strategy, we are able to estimate accurate, metrically-scaled extrinsics from unlabeled image sequences, without expensive optimization or bundle adjustment.
Contrary to standard extrinsic calibration pipelines, which require specific tools (including \textit{calibration targets}~\cite{Chen2022extrinsic}) or time-consuming computation (bundle adjustment~\cite{Lin2020ECCV}), we use photometric consistency as a training objective. As a result, we can self-calibrate a multi-camera system without depending on the supervision of 3D sensors or any manual labor.


Experimental results show that our approach produces 
more robust calibration on a variety of real-world driving scenes compared to a standard structure-from-motion tool (COLMAP~\cite{schonberger2016structure}). 
Moreover, our results also reveal specific difficulties and limitations in each sequence, both quantitatively and qualitatively. 
Finally, we also provide evidence that our predicted extrinsics improves depth estimation performance via the joint self-supervised optimization of extrinsics alongside depth and ego-motion.

In summary, we propose \textbf{\Acronym} (Self-Supervised Extrinsic Self-Calibration), and our contributions are as follows:

\begin{itemize}
    \item We introduce a \textbf{novel camera extrinsics self-calibration technique} for wide-baseline multi-camera systems utilizing the self-supervised monocular depth estimation paradigm (See Figure~\ref{fig:teaser} for overview).

    \item We thoroughly investigate our self-calibration performance on benchmark datasets (DDAD~\cite{packnet} and KITTI~\cite{geiger2012we}), and show its performance and limitations in different scenarios. 

    \item We demonstrate that self-supervised training of monocular depth and pose networks jointly with \textbf{extrinsics self-calibration improves depth estimation} performance on KITTI~\cite{geiger2012we} and DDAD~\cite{packnet}.

\end{itemize}

%% file: sections/related.tex
\noindent\textbf{Multi-Camera Wide-baseline Depth Estimation.}
Early works on learning-based depth estimation from a single image were fully supervised~\cite{eigen2014depth,laina2016supervised}, requiring ground truth depth maps for training. After that, the need for large amounts of high-quality annotated data led research efforts towards self-supervised depth estimation~\cite{zhou2017unsupervised,godard2017unsupervised}. Zhou~\emph{et al.}~\cite{zhou2017unsupervised} formulated this as a joint optimization problem of depth and ego-motion estimation via photometric consistency using \emph{STN}~\cite{jaderberg2015spatial}. Godard~\emph{et al.}~\cite{monodepth2} developed strategies to alleviate some of the inherent problems that come from structure-from-motion learning, while other works focused on architecture design~\cite{packnet}, presence of dynamic objects~\cite{hui2022recurrentmono}, and uncertainty estimation~\cite{Hornauer2022GradientbasedUF}.
Importantly, all these works were limited to the \textit{monocular} setting, in which the field of view is limited to be that of a single camera.

To achieve self-supervised wide-baseline depth estimation, FSM~\cite{fsm20222} proposed the use of spatial-temporal photometric and pose consistency across cameras. Differently from other methods, FSM~\cite{fsm20222} does not require large amounts of overlap between cameras or 360\textdegree\ images~\cite{manuel2022360mono}, which allows for more flexibility in the mechanical design of the system and the integration of other vision-based perceptual modules. 
Subsequently, other methods were proposed to improve multi-camera depth estimation performance, in terms of architecture, method of prediction, and loss function design~\cite{kim2022selfsupervised,xu2022multicollab,wei2022surround}.  
However, all these works assume well-calibrated multi-camera extrinsics, which is a strong assumption in many practical scenarios~\cite{Chen2022extrinsic}.
Thus, a reliable and efficient method is required in order to fully realize the benefit of calibrated extrinsic parameters in multi-camera depth estimation. 

\noindent\textbf{Camera Calibration via Photometric Consistency}
Traditional approaches to camera calibration are \emph{target-based}: images of well-known patterns such as AprilTags~\cite{olson2011apriltag} and checkerboards are captured and used to solve a correspondence matching problem by optimizing the relationship between camera parameters and observations. 
Even though recent open-source tools have facilitated this process and made it more efficient~\cite{rehder2016kalibr,Rameau2022MCCalib}, the need for tools and a pre-designed structure makes it difficult to run the calibration procedure without access to the system itself~\cite{koo2021extrinsics}. 
Recently, a \emph{targetless} strategy using image features from the surrounding environment was proposed in~\cite{Lin2020ECCV,Hu2022Calib}. By using COLMAP~\cite{schonberger2016structure} and additional optimization, Liu~\emph{et al.} achieved calibration from arbitrary images independently of the initial guess or the use of tools or/and pre-structured environments. 
However, this is still (1) a time-consuming process, that requires feature extraction and matching; and (2) it frequently fails to achieve proper convergence~\cite{fsm20222}, thus making it difficult to efficiently and robustly apply this method to large-scale real-world datasets. 
Motivated by this fact, we propose a novel deep learning-based strategy that does not require explicit feature extraction and matching (which is usually sparse and non-differentiable), 
and achieves robust convergence for a variety of domains
by utilizing dense and more informative geometric constraints~\cite{jeong2021scnerf,zhang2022relpose}. 
Especially, using the photometric consistency between two images as a training objective removes the dependency on additional pose labels for supervision~\cite{zhang2022relpose} or 3D sensory input like LiDAR~\cite{Hu2022Calib}.
Furthermore, as our learned depth and ego-motion networks can be directly applied to other camera configurations, we can simultaneously and efficiently calibrate multi-camera systems.

Photometric consistency provides a strong training signal that can be used as proxy supervision to learn several different geometry-based tasks.
In particular, by optimizing image reconstruction ~\cite{jeong2021scnerf,yen2020inerf,lin2021barf} or depth estimation ~\cite{gordon2019depth,vasiljevic2020neural,tri-selfcalibration} networks, previous works have shown that it is possible also to learn intrinsic and/or extrinsic camera parameters. Since mitigation of learning difficulties in a dynamic environment~\cite{monodepth2} and relatively fast inference time is attractive for efficient calibration in various real-world environments, we chose the latter group, which explicitly learns depth (and the corresponding ego-motion) and camera parameters. 
Our method is most similar to the self-supervised self-calibration work from~\cite{tri-selfcalibration}, where the authors demonstrated that camera intrinsics could be jointly learned alongside a depth and pose networks by minimizing the photometric loss in a single-camera setting.
Here, we extend the self-calibration of intrinsics to the multi-camera setting, then show that it is possible to jointly learn intrinsic \textit{and extrinsic} parameters alongside depth and pose, improving depth estimation in a multi-camera setting.



%% file: sections/methodology.tex
We now describe \Acronym, our proposed framework to learn depth, ego-motion, and extrinsic camera parameters jointly in a self-supervised manner. 
First, we formulate the problem and summarize the standard method of self-supervised monocular depth and ego-motion learning.
Next, we introduce our extensions towards multi-camera self-supervised extrinsic self-calibration, describing the challenges of this new setting.
Finally, we present a curriculum learning method that leverages the learned extrinsics to further improve the depth and pose estimates.
We assume a set of cameras rigidly attached to a vehicle, with arbitrary relative position and overlap between pairs. We also assume the instantaneous velocity of the vehicle and whether each camera is pointing backward or not.

\subsection{Self-Supervised Depth and Ego-motion Training}
\label{sec:mono}
Self-supervised depth and ego-motion learning is defined as the joint optimization of a \emph{depth} network, which maps a target image $I_t$ to a depth map $\hat{D}_{t}$, as well as a \emph{pose} network, that predicts the relative transformation  $\hat{\mathbf{X}}^{t \to c} = \begin{psmallmatrix}\mathbf{\hat{R}^{t\to c}} & \mathbf{\hat{t}^{t\to c}}\\ \mathbf{0} & \mathbf{1}\end{psmallmatrix} \in \text{SE(3)}$ from a context $c$ to target $t$ frames.
The synthesized target image $\hat{I}_t$ is obtained via reprojection~\cite{zhou2017unsupervised} using $\hat{D}_{t}$, $\hat{\mathbf{X}}^{t \to c}$, camera intrinsics $\textbf{K}_t$, and the context image $I_c$. The photometric reprojection error is then calculated as:
\begin{equation}
\mathcal{L}_{p}(I_t,\hat{I_t}) = \alpha~\frac{1 - \text{SSIM}(I_t,\hat{I_t})}{2} + (1-\alpha)~\| I_t - \hat{I_t} \|
\label{eq:photo_mono}
\end{equation}
Here, $\mathcal{L}_{p}(I_t,\hat{I_t})$ is the photometric reprojection loss~\cite{zhou2017unsupervised,godard2017unsupervised}, and the SSIM term measures the structure similarity of two images~\cite{wang2004image}. 
This reprojection error provides the photometric consistency constraint we use as the self-supervised training objective.
In addition, to promote smoothness in the predicted depth map, a regularization term is added to the above reprojection error~\cite{packnet, godard2017unsupervised}:
\begin{equation}
    \mathcal{L}_{s}(\hat{D_t}) = \vert \delta_x\hat{D_t} \vert e^{ - \vert \delta_{x}I_{t} \vert } + \vert \delta_y\hat{D_t} \vert e^{ - \vert \delta_{y}I_{t} \vert }
    \label{eq:smoothness}
\end{equation}

To address our multi-camera setup, we follow~\cite{fsm20222} and define spatio-temporal constraints via photometric consistency between reprojected camera images across the spatial (i.e. between different cameras) as well as temporal axes (i.e. between different timesteps).


\subsection{Self-Supervised Extrinsic Self-Calibration}
\label{sec:pcc++}
In the discussion below, we index cameras by subscripts and temporal indices (video frames) by superscripts.  Also, \textit{pose} refers to the motion of the vehicle, while \textit{extrinsics} refers to the poses of cameras on the rig with respect to a vehicle coordinate frame. Note that we chose Euler angles as a representation
 for extrinsics because of their simplicity, but other parameterizations are possible.

First, we will review the pose consistency loss proposed in~\cite{fsm20222} which assumes calibrated extrinsic, and then extend it to our self-calibration setting.
Given a camera $i$, the pose network predicts its transformation $\hat{\mathbf{X}}^{t \to \tau}_{i}$ from the current frame $t$ to a temporally adjacent frame $\tau$ (usually $\tau \in \{t-1,t+1\}$). Assuming that all cameras are rigidly attached to the vehicle, their motion should be consistent when observed from the vehicle coordinate system, i.e. the motion of the vehicle estimated from $t \to \tau$ in camera $i$ should be the same as the motion observed in camera $j$ (given that the cameras are rigidly attached).
Given the known extrinsics $\mathbf{X}_{i}$ and $\mathbf{X}_{j}$, respectively, we can transform the vehicle pose estimated using the pose network predictions from camera $i$ ($\hat{\mathbf{X}}^{t \to \tau}_{i}$) into camera $j$ using:
\begin{equation}
    \tilde{\mathbf{X}}_{i}^{t \to \tau} 
    = \mathbf{X}_j^{-1}\mathbf{X}_i
    \hat{\mathbf{X}}_{i}^{t \to \tau}
    \mathbf{X}_i^{-1}\mathbf{X}_j
    \label{eq:pcc_mat}
\end{equation}
The above transformation can be used to transform pose predictions from all cameras into the coordinates of camera $1$, which is the label we give to the forward-facing camera. 
Given that there is inconsistency in the pose predictions, FSM~\cite{fsm20222} defines a pose consistency loss as the weighted sum of pair-wise translation and rotation errors: 
\begin{equation}
    \begin{split}
    \mathcal{L}_{pcc}(
        \hat{\mathbf{X}}^{t \to \tau}_{1},
        \tilde{\mathbf{X}}_{j}^{t \to \tau}
        ) 
    = \sum_{\tau} 
      \sum_{j=2}^{N} \lbrace
      \alpha_{t} \mathcal{L}_{t}(
        \hat{\mathbf{t}}^{t \to \tau}_{1}, 
        \tilde{\mathbf{t}}^{t \to \tau}_{j}
        ) \\
     + \alpha_{r} \mathcal{L}_{R}(\hat{\mathbf{R}}^{t \to \tau}_{1}, \tilde{\mathbf{R}}^{t \to \tau}_{j})  \rbrace
    \end{split}
    \label{eq:pcc_loss}
\end{equation}
where
$\hat{\mathbf{X}}^{t \to \tau}_j = \begin{psmallmatrix}\mathbf{\hat{R}}_{j}^{t \to \tau} & \mathbf{\hat{t}}_{j}^{t \to \tau}\\ \mathbf{0} & \mathbf{1}\end{psmallmatrix} \in \text{SE(3)}$.

In the above equation, $N$ is the number of cameras, and $\alpha_{t}$ and $\alpha_{R}$ are respectively the translation and rotation weighting parameters. In~\cite{fsm20222}, the squared difference is applied for both $\mathcal{L}_{t}$ and $\mathcal{L}_{R}$ after converting the rotation matrix to Euler angles. 
For the fully-calibrated setting in~\cite{fsm20222}, this consistency constraint is used to improve the pose as well as depth network predictions.

\textbf{Pose Consistency for Extrinsics Learning:}
Here extrinsics refer to the poses of the cameras on the rig with respect to the vehicle coordinate frame, which in our experiments is chosen as the origin of the front camera. Therefore, $\mathbf{X}_1$ is $\mathbf{I}$.
Note that the camera extrinsics $\mathbf{X}_i$ are present in Equation~\ref{eq:pcc_mat} and thus in Equation~\ref{eq:pcc_loss}; this can be used as a differentiable constraint on the extrinsic parameters, and we use it to guide extrinsics learning.
However, naively learning extrinsics from scratch, guided by Equation~\ref{eq:pcc_loss}, causes extrinsics learning to fail because of a property of the pose network: the predicted rotation $\mathbf{\hat{R}}^{t\to \tau}$ is close to the identity matrix when the images are densely sampled and the camera moves in a straight line, which is common in driving scenarios.
Next, we describe our initialization technique to alleviate this and enable joint depth, pose, and extrinsics estimation.

Consider this early training scenario where the predicted pose $\hat{\mathbf{R}}^{t \to \tau}_{j} \approx \mathbf{I}$ and 
$\hat{\mathbf{X}}_j = \begin{psmallmatrix}\mathbf{\hat{R}}_{j} & \mathbf{\hat{t}}_j\\ \mathbf{0} & \mathbf{1}\end{psmallmatrix} \in \text{SE(3)}$, plugging Equation~\ref{eq:pcc_mat} into Equation~\ref{eq:pcc_loss} will look like:
\begin{equation}
    \begin{gathered}
    \mathcal{L}_{t}(\hat{\mathbf{t}}^{t \to \tau}_{1}, \tilde{\mathbf{t}}^{t \to \tau}_{j})
    \approx   
    \mathcal{L}_{t}(
        \hat{\mathbf{t}}^{t \to \tau}_{1}, 
        \mathbf{\hat{R}}_{j}^{-1}\hat{\mathbf{t}}^{t \to \tau}_{j}
        )
    \\
    \mathcal{L}_{R}(\hat{\mathbf{R}}^{t \to \tau}_{1}, \tilde{\mathbf{R}}^{t \to \tau}_{j})
    \approx   
    \mathcal{L}_{R}(
        \mathbf{I}, \mathbf{\hat{R}}_{j}^{-1}\mathbf{\hat{R}}_{j}
        )
    \end{gathered} 
    \label{eq:pcc_approx}
\end{equation}
Here the rotation matrices are represented as Euler angles such that $\mathbf{\hat{R}}(\theta, \phi, \psi) = \mathbf{\hat{R}}_{x}(\theta)\mathbf{\hat{R}}_{y}(\phi)\mathbf{\hat{R}}_{z}(\psi)$. 
If the ego-motion in the dataset is only composed of forward or backward motion (giving $\hat{\mathbf{t}}^{t \to \tau}_{1} \approx  (0,0,const.)^T$ ), and the camera $j$ is looking backwards relatively to camera $i$ (which gives $\hat{\mathbf{t}}^{t \to \tau}_{j} \approx  (0,0,-const.)^T$ ), the above approximation shows that there is no useful gradient for $\psi$. Therefore, we set $(\theta, \phi, \psi) = (\pi, 0, \pi)$ only for the initialization of the back-looking camera, and set $(\theta, \phi, \psi) = \mathbf{0}$ for the others. 

Note that the approximation in Equation \ref{eq:pcc_approx} also reveals another limitation of our training procedure: it is difficult to get the gradient of the translation vector $\mathbf{\hat{t}}_j$. 
Thus, we learn the camera translations in a sequential manner, first estimating the rotation 
and then learning the translation (followed by end-to-end finetuning), please refer to Figure~\ref{fig:flow} and Section \ref{sec:curriculum} for more details.


\input{tables/curriculum_description}
\input{figures/learning_flow.tex}

\subsection{Curriculum Learning}
\label{sec:curriculum}
Jointly learning depth, pose, and extrinsics from scratch is a highly underconstrained problem, and empirically we have observed that a naive approach fails to converge in the self-supervised setting. 
Thus, we propose a \emph{curriculum schedule} composed of different learning modules (depth, ego-motion, and extrinsic parameters) and objectives to minimize (photometric loss and pose consistency loss), designed to enable self-supervised learning in this novel setting. Table~\ref{tab:curriculum_description} summarizes the different stages of our proposed curriculum learning schedule and Figure~\ref{fig:flow} shows the extrinsics update flow. Below we describe each stage in detail and quantify their impact in the experiments section. 

\subsubsection{Monodepth Pretraining}
We train our depth and pose estimators following monodepth2~\cite{monodepth2}. 
Removing the scale ambiguity in depth and ego-motion learning is an important factor in our self-calibration strategy, as a way to make the target extrinsics metrically scaled. 
To inject metric scale without any calibration procedure, similar to Kumar~\emph{et al.}~\cite{fisheye2020}, we normalize the output of the pose network and multiply it by the magnitude of the relative translation. The relative translation is obtained using ground-truth velocity, defined as the instantaneous velocity divided by the image sampling frequency. 


\subsubsection{Rotation Estimation}
As mentioned in~\ref{sec:pcc++}, the purpose of this stage is the robust initialization of extrinsic rotation vectors, rather than translation vectors. 
With a frozen pretrained pose network, we optimize the extrinsic rotation parameters using only the pose consistency loss (Equation~\ref{eq:pcc_loss}).

\subsubsection{Extrinsic Estimation}
In this stage, all extrinsic parameters and the pose network are jointly optimized using both the photometric and pose consistency losses.

\subsubsection{End-to-end Training}
Finally, the entire architecture (including the depth network) is jointly optimized with extrinsics parameters using the photometric and pose consistency losses. 


%% file: tables/curriculum_description.tex
\captionsetup[table]{skip=6pt}

\begin{table}[t!]
\vspace{+3mm}
\renewcommand{\arraystretch}{1.00}
\centering
{
\small
\setlength{\tabcolsep}{0.25em}
\begin{tabular}{l|ccc|cc}

\toprule
\multirow{2}{*}{\textbf{Stage}}
& \multicolumn{3}{c|}{Optimization}
& \multicolumn{2}{c}{Loss} 
\\
\cmidrule{2-6}
& \textit{depth}
& \textit{ego-motion}
& \textit{extrinsics}
& \textit{Photo}
& \textit{Pose}
\\
\midrule

\emph{Monodepth Pretraining}
& \checkmark 
& \checkmark 
& -
& \checkmark 
& \checkmark 
\\
\emph{Rotation Estimation }
& -
& \emph{Fix}
& \checkmark 
& -
& \checkmark 
\\
\emph{Extrinsic Estimation}
& \emph{Fix}
& \checkmark 
& \checkmark 
& \checkmark 
& \checkmark  
\\
\emph{End-to-end Training}
& \checkmark 
& \checkmark 
& \checkmark  
& \checkmark 
& \checkmark  
\\
\bottomrule
\end{tabular}
}
\caption{
\textbf{Optimization and losses during extrinsic learning.} At each stage, different models are optimized with different losses. The symbol \checkmark indicates a component that is being optimized at that stage (and which loss is used), and \emph{Fix} indicates a component that is fixed (not optimized). 
}
\label{tab:curriculum_description}
\end{table}

%% file: figures/learning_flow.tex
\begin{figure}[t!] 
    \centering
    \includegraphics[width=0.40\textwidth]{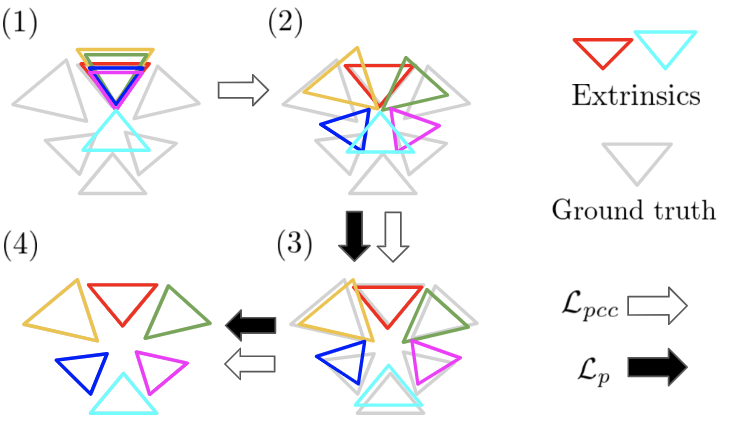}
    \caption{\textbf{An example of curriculum learning of extrinsics.} (1) Extrinsics are initialized after \emph{Monodepth Pretraining}, (2) rotation parameters are primarily optimized in the \textit{Rotation Estimation} stage (via the \emph{Pose Consistency} loss), (3) all extrinsic parameters are updated in the \emph{Extrinsic Estimation} stage (via the \emph{Photometric} loss), and (4) the final extrinsics are predicted in the \emph{End-to-end training} stage. }
    \label{fig:flow}
    \vspace{-3mm}
    \end{figure}

%% file: sections/experiments.tex

We compare \Acronym to strong baselines for depth estimation as well as camera calibration. We aim to show (1) that \Acronym achieves competitive extrinsic calibration in a challenging multi-camera setting where each camera shares an arbitrary amount of field of view overlap as well as in the simplest multi-camera setup of a stereo pair; and (2) that the extrinsic calibration learned by \Acronym in turn helps improve monocular depth estimation in the self-supervised setting.


\input{figures/ddad_calib_many_sequences}

\subsection{Datasets}

\textbf{DDAD \cite{packnet}}. 
DDAD (Dense Depth for Autonomous Driving) is a dataset with images recorded by $6$ cameras mounted around the vehicle, forming a $360^o$ field of view. It has a total of $12650$ training samples (composed of $150$ sequences each with $50$ or $100$ images) and $3950$ validation samples (composed of $50$ sequences each with $50$ or $100$ images). All cameras are synchronized at $10$ Hz,  which facilitates self-supervised learning in a multi-camera setting. In all our experiments, images are downsampled to a resolution of $640 \times 384$, and evaluated at distances of up to 200 meters. Furthermore, because the training sequences were collected using different vehicles in different cities, we learn instead per-sequence extrinsic parameters, as a way to investigate the performance of our self-calibration method in different scenarios. As reported in ~\cite{fsm20222}, the DDAD dataset has several areas with self-occlusion from the vehicle body, which can harm monocular self-supervised learning.  Thus, we use the same self-occlusion masking protocol for training.  The ground-truth depth maps are used only for evaluation.
In contrast to temporally adjacent image pairs that share $30-95\%$ field of view, the overlap area of spatial- and spatiotemporal-image pairs only shares $5-30\%$. 

\textbf{KITTI \cite{geiger2012we}}.
The KITTI dataset is the standard benchmark for depth and ego-motion estimation. 
We train our model on the \textit{Eigen} train split \cite{zhou2017unsupervised} containing $22,600$ images per camera and validate on the test split containing $697$ images. Images are reshaped to $640 \times 192$, and evaluated at distances up to 80m with the \emph{garg} crop~\cite{zhou2017unsupervised}. Extrinsics are learned according to timestamp date, so five configurations are trained from \emph{2011\_09\_26} to \emph{2011\_10\_03}. 
A target image and corresponding image always overlap about $50-70\%$. 

\subsection{Implementation Details}

Our models were implemented using PyTorch~\cite{paszke2017automatic} and trained across eight NVIDIA A100 GPUs of $80$ GB per device.
Previous and subsequent frames are used as temporal context. Color jittering and left-right image flips are applied as data augmentation for the depth network.

For the \emph{Monodepth Pretraining} stage, we use the Adam optimizer~\cite{kingma2014adam} with learning rate $lr = 8 \times 10^{-5}$ for both the depth and pose networks, a batch size $b=4$, and train for a total of $15$ epochs. 
Note that this step is done using pretrained weights, obtained by training standard self-supervised monocular depth and ego-motion networks considering each camera independently, without weak velocity supervision. Empirically, we found that the direct learning of scaled depth and ego-motion creates instability in the training pipeline.

After pretraining the scale-aware model, we begin the extrinsic learning stage with the following parameters: $b=3$ per GPU, the SGD optimizer is used with $lr=0.1$ for the \emph{Rotation Estimation} stage, the Adam optimizer is used for the \emph{Extrinsic Estimation} stage with the $lr=2 \times 10^{-4}$ for the pose network and $lr=1 \times 10^{-3}$ for the extrinsics parameters, and the Adam optimizer is also used for the last joint optimization with $1 \times 10^{-5}$ for the depth and pose networks and $1 \times 10^{-4}$ for the extrinsic parameters. The number of epochs is set to 10 for the \emph{Rotation Estimation} and \emph{Extrinsic Estimation} stages, and 5 for the \emph{End-to-end Training} stage.
For the pose consistency, the translation term is formulated following~\cite{fsm20222}, and the geodesic loss~\cite{hempel2022sixd} is used as the rotation term. 
We set $\alpha_{t} = \alpha_{R}=1.0$ in all experiments.

For the network architectures, following~\cite{monodepth2}, we used a ResNet18-based depth and pose networks. Additionally, we introduce ``per-camera" extrinsic parameters, which are optimized during training. We evaluate \Acronym on datasets which consist of \emph{sequences} collected by vehicles that have the same sensory setup, but slight variations in their extrinsic parameters across sequences due to calibration differences as well as temporal variations. To account for this, we estimate one set of per-camera parameters over the whole dataset, as well as a per-camera residual which is optimized for each sequence. 



\input{tables/vs_colmap_full.tex}

\subsection{Multi-camera extrinsic self-calibration on DDAD}

\textbf{Extrinsics Calibration Performance. } Our results are summarized in Table~\ref{tab:full_colmap}, and we note that \Acronym achieves on average $0.16$m translation error and below $1\degree$ rotation error for extrinsic calibration. We compare \Acronym with COLMAP~\cite{schonberger2016structure}, a widely used structure-from-motion library. 

Since COLMAP predicts intrinsics as well as extrinsics, we present a version of our method, denoted by \AcronymInt, where we also learn the intrinsics for all the cameras in a self-supervised manner, following the approach described in Fang~\emph{et al.}~\cite{tri-selfcalibration}: the ground-truth intrinsic parameters are replaced with a learnable parameter vector which is optimized jointly with the other networks. The intrinsics are initialized using the image resolution (i.e. $H/2, W/2$), following~\cite{tri-selfcalibration}, and we learn one parameter vector for each camera which is optimized across the entire dataset. 
The Adam optimizer with $lr=1 \times 10^{-2}$ is applied to these parameters in all stages except \emph{Rotation Estimation}. We note that this version of our model also achieves competitive results even in this very challenging setting of unknown intrinsics and extrinsics.

We run COLMAP separately on each DDAD sequence, and we metrically scale COLMAP predictions by adjusting their magnitude based on ground-truth relative translation, to ensure a fair comparison with our method. In all experiments, we run bundle adjustment with the \emph{rig\_bundle\_adjuster} option and set each camera model as \emph{pinhole}, to account for the fact that all cameras are rigidly attached to the vehicle and can be modeled using a pinhole geometry.


\input{tables/table_metric_ddad2}

Table~\ref{tab:full_colmap} shows a quantitative comparison between the two methods. 
We observed that COLMAP did not converge in all sequences. Furthermore, even when it converged, the translation error produced by COLMAP was very large (up to a few meters) in several sequences. Therefore, for a more nuanced comparison, we also report COLMAP results on the sequences where convergence was reached (for a total of $97$ sequences, marked as COLMAP\textsuperscript{*}), and those in which the translation error $t$ is smaller than $1$m (for a total of $89$ sequences, marked as COLMAP\textsuperscript{**}).
We note that our method outperforms COLMAP\textsuperscript{**} on translation error and is competitive on orientation error. Note that both \Acronym(\emph{I}) and \Acronym were able to converge in all sequences from the same initialization, as an indication that it is more robust to different environment conditions. Moreover, the total training time of our method is $\sim12$ hours for the whole dataset, while COLMAP required $150\sim$ hours ($\sim 1$ hour per sequence if it converged, otherwise more time is required, up to 4 hours).

A qualitative comparison between our calibrated camera extrinsics and COLMAP is shown in Figure~\ref{fig:ddad_calib_many}, alongside our predicted point clouds.
These sequences are from significantly different settings: (a) a street scene at low speeds with mostly parked cars, and (b) a highway scene at high speeds with many dynamic objects. The latter is very challenging for COLMAP, as motion blur and the presence of dynamic objects makes it difficult to perform feature matching. Even though our method also degrades in this challenging setting, compared to the simpler former setting, we are still able to reasonably estimate camera extrinsics and depth maps.

\textbf{Depth Estimation Evaluation. }
Next, we evaluate \Acronym on the task of depth estimation, to answer the question of whether our self-calibrated extrinsics can be used as proxy to ground-truth extrinsics and improve depth estimation performance by leveraging cross-camera constraints. 
These results are reported in Table~\ref{tab:depth_ddad}, including variations of our depth network trained in different conditions. In \emph{Monodepth+v} we directly use the backbone from the \emph{Monodepth Pretraining} stage. \Acronym$\dagger$ is trained using the same curriculum and number of epochs as \Acronym but without optimizing extrinsics; we include this baseline to highlight the fact that learning intrinsics via our method leads to improved depth performance, and that the increase is not due to additional training of the depth and ego-motion networks. \Acronym and \Acronym$\dagger$ are trained $7$ times using randomly generated seeds and the same hyperparameters, and we also report the mean of these runs. Finally, we also include FSM~\cite{fsm20222} results obtained using the same ResNet18-based backbone.


These result shows that \Acronym, which optimizes extrinsics as well as depth and ego-motion networks, 
improves depth estimation performance compared to the baseline of \mbox{\Acronym$\dagger$}.
Furthermore, these results also show that \Acronym is competitive with methods that use ground-truth extrinsics, such as FSM~\cite{fsm20222}.
In fact, our model outperforms FSM in several cameras (especially \emph{B.Right}) although we use the same depth backbones. We hypothesize this is due to a slight miscalibration in the provided ground-truth extrinsics.
In summary, our results are evidence that the learned extrinsics generated by \Acronym  can contribute to improvements in depth estimation without requiring additional information or explicit cross-camera calibration~\cite{fsm20222}.
\input{figures/stop_vehicle_000032}

\input{tables/curriculum_learning_ddad}

\textbf{Effect of Curriculum Learning. }
In Table~\ref{tab:curriculum_l}, we investigate the effects of our proposed curriculum strategy on the DDAD dataset to show that it enables the learning of extrinsic parameters without task-specific prior knowledge. In this process, the first stage (\emph{Rotation Estimation}) promotes the proper initialization of camera orientation from initial values. Afterwards, the second stage (\emph{Extrinsic Estimation}) significantly reduces translation error while also improving rotation. Finally, the joint optimization further improves accuracy, including depth estimation performance.


By ablating both the \emph{Rotation Estimation} (see \emph{Extrinsic Est.} $-$ 2) and \emph{Extrinsic Estimation} (see \emph{E2E} $-$ 3) stages we show that \Acronym effectively decreases translation and rotation error at each training stage.

\subsection{Stereo-camera extrinsic self-calibration on KITTI}

Table~\ref{tab:stereo_calib} shows the estimated error of our predicted extrinsic parameters in each learning stage, for the KITTI stereo setup. Even in this ``simplest" calibration case, our proposed method improves performance through the proposed stages of our curriculum, achieving a final error of 0.018[m]. 
The depth estimation performance of the model is shown in Table~\ref{tab:stereo}. 
\Acronym achieves better results on all metrics compared to the baseline of \Acronym$\dagger$ (which does not use predicted extrinsics). In addition, \Acronym is competitive with the current state-of-the-art, even though it relaxes the stereo-camera assumption and does not have access to information about the stereo baseline.

\input{tables/calibraiton_stereo}
\input{tables/table_stereo}

\section{Limitations}

As \Acronym relies on the self-supervised learning of depth and ego-motion, there are some settings that are challenging, due to limitations in the photometric loss used as training objective. For example, \Acronym fails to properly calibrate  sequences in which there are frequent complete stops, e.g. as shown in Figure~\ref{fig:seq52} (\emph{seq:000032}). Since self-supervised depth and pose networks rely on camera ego-motion, these sequences do not provide any meaningful source of supervision for our self-supervised training procedure, leading to a failure in calibration. 

%% file: figures/ddad_calib_many_sequences.tex
\begin{figure*}[t!]
\centering
\vspace{2mm}

    \begin{tabular}{cc}
        \begin{minipage}{.80\textwidth}
            \subfloat[\emph{seq:000052} A street scene at low speeds with mostly parked cars. Both methods achieve good results.]{ 
            \includegraphics[width=0.28\textwidth]{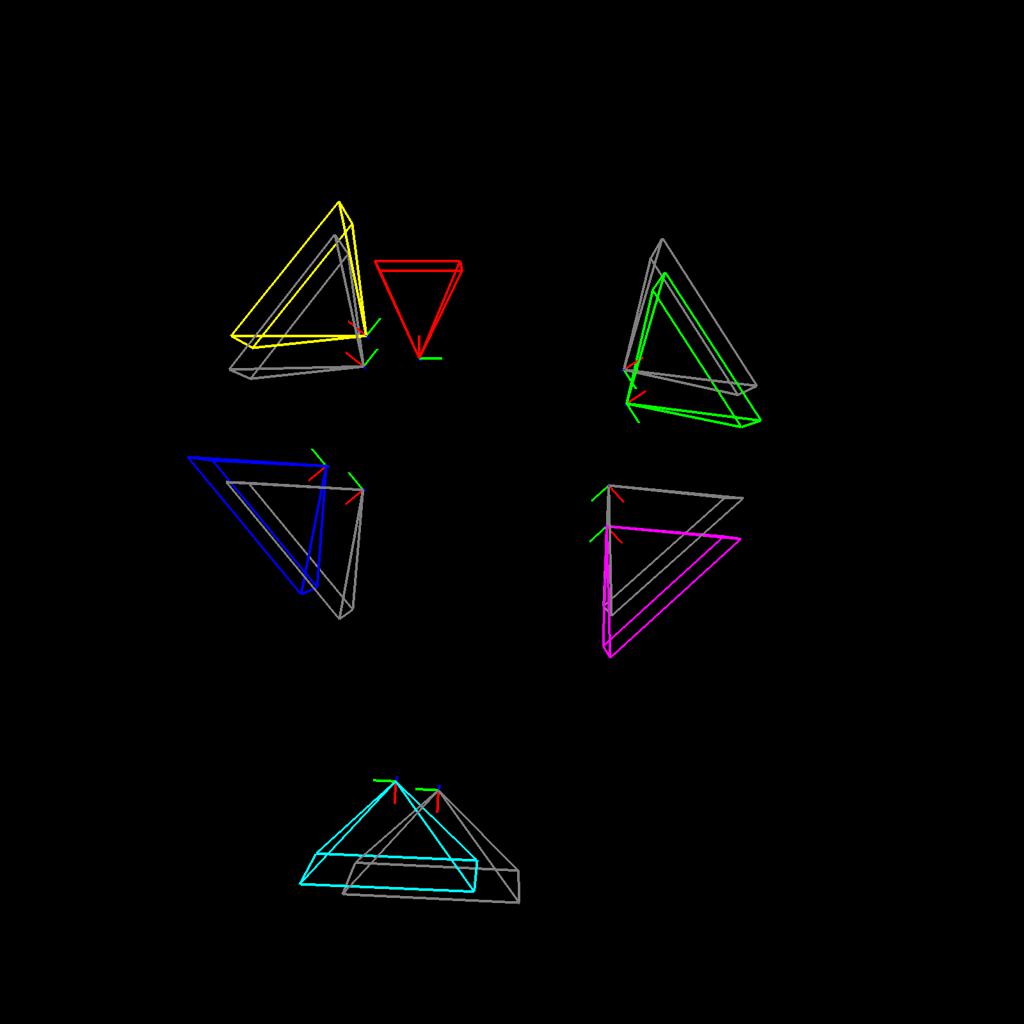}
            \includegraphics[width=0.28\textwidth]{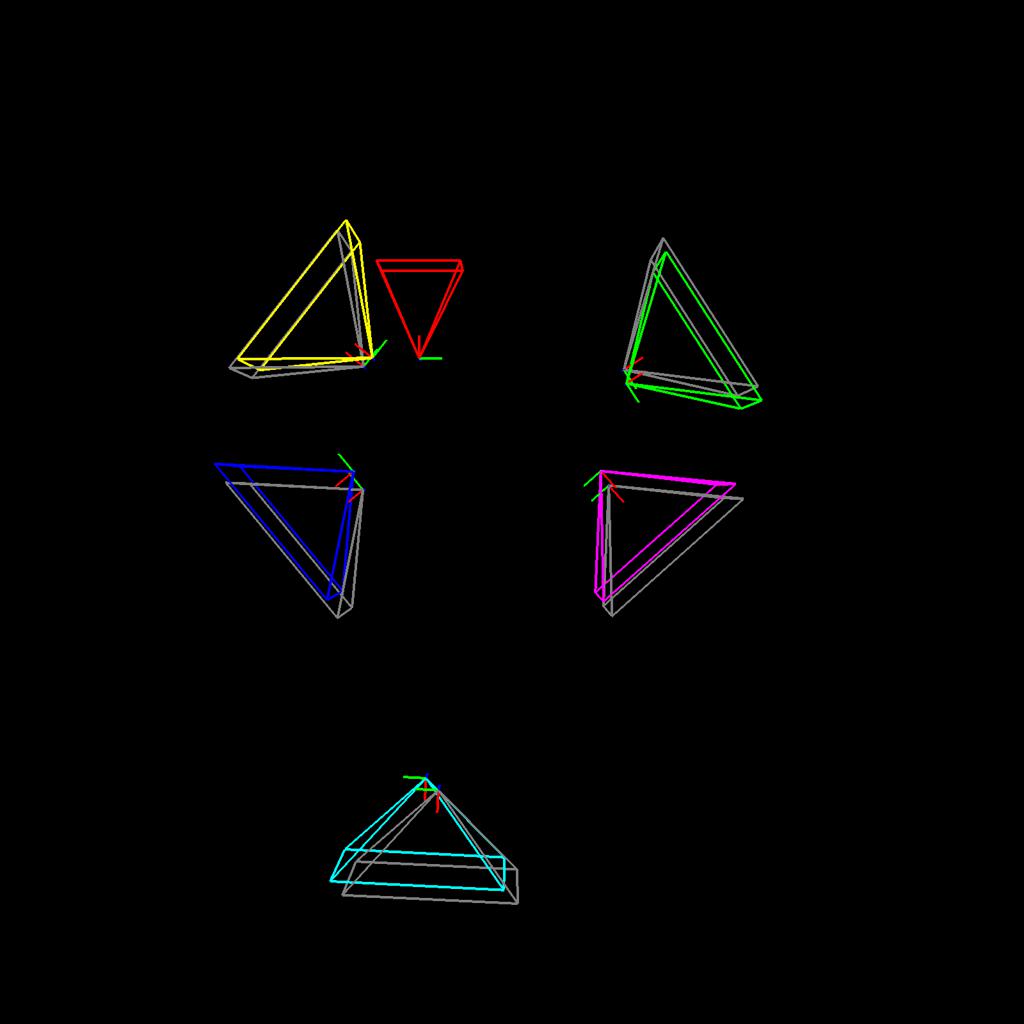}
            \includegraphics[width=0.28\textwidth]{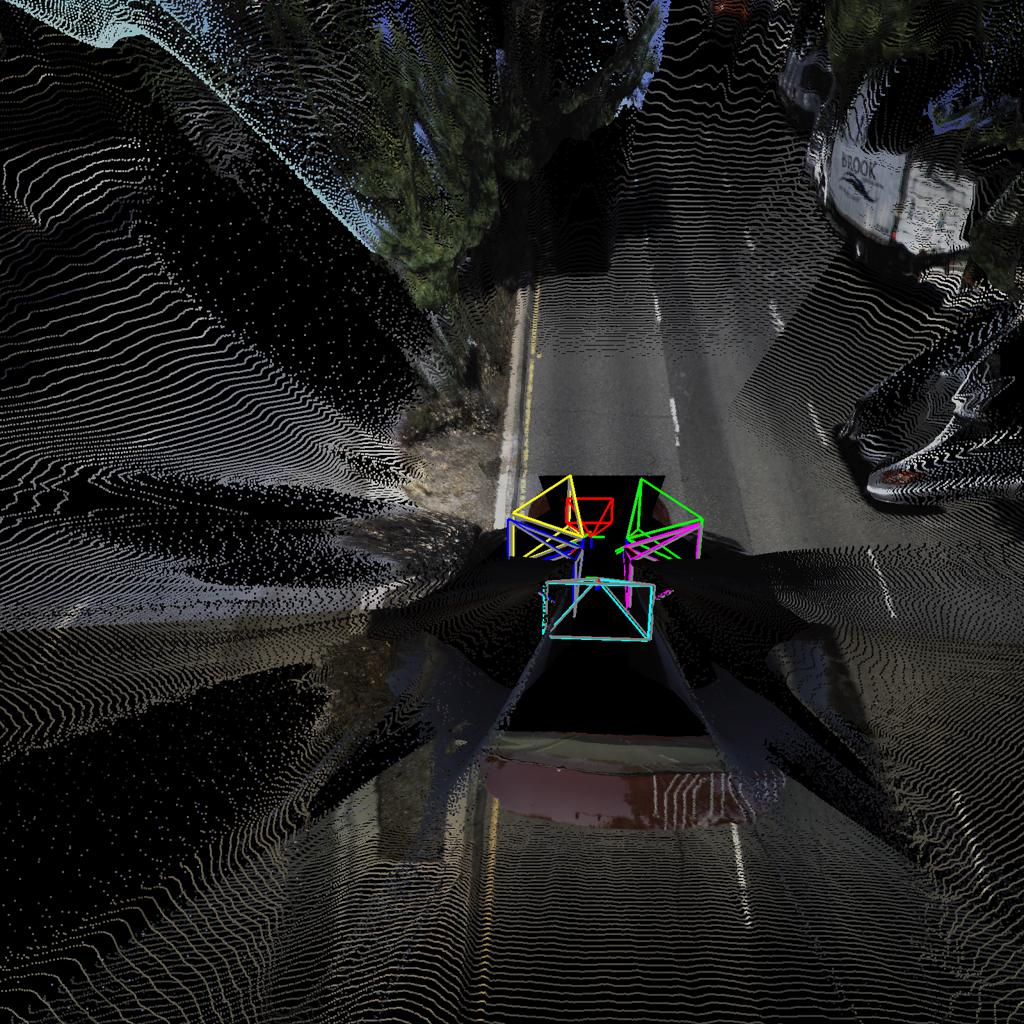}
            }
            \\
            \subfloat[\emph{seq:000016}: A highway scene at high speeds with many dynamic objects. COLMAP fails while \Acronym still achieves competitive results.]{ 
            \includegraphics[width=0.28\textwidth]{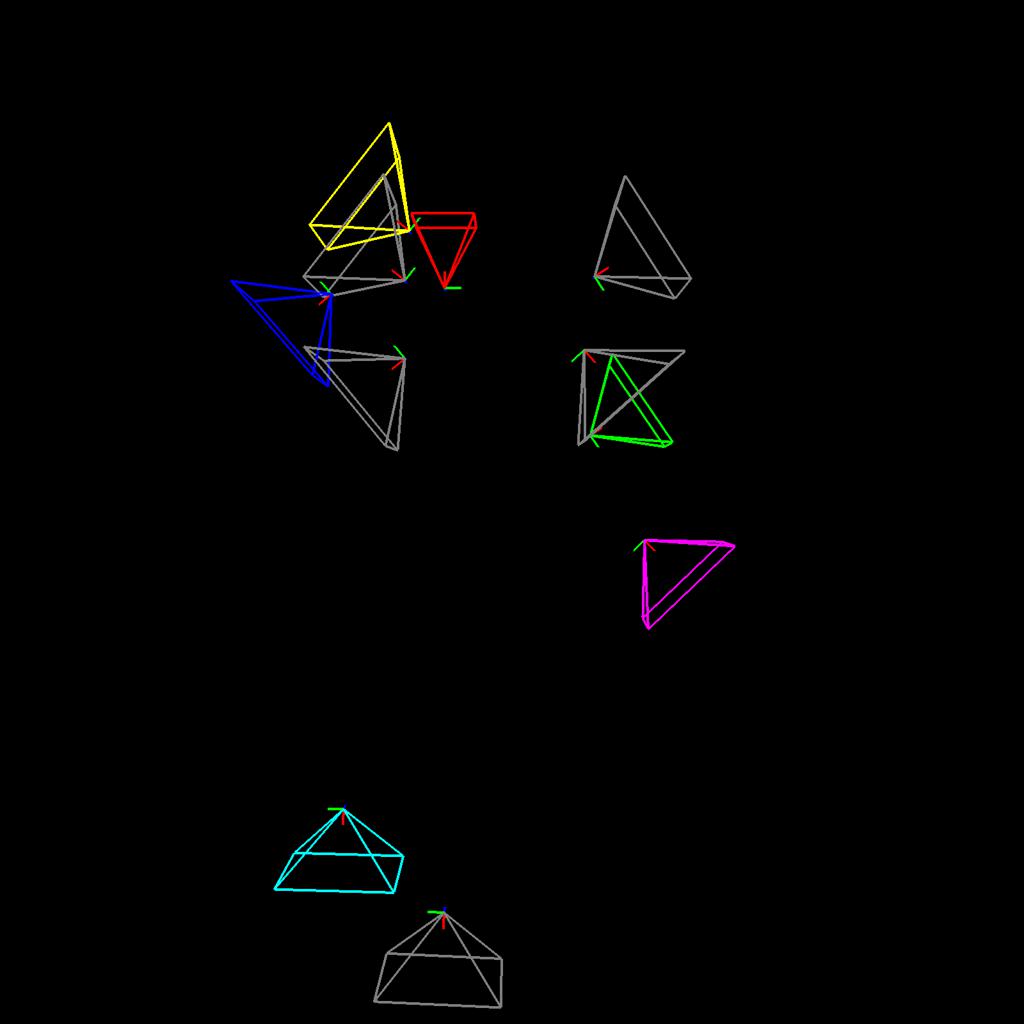}
            \includegraphics[width=0.28\textwidth]{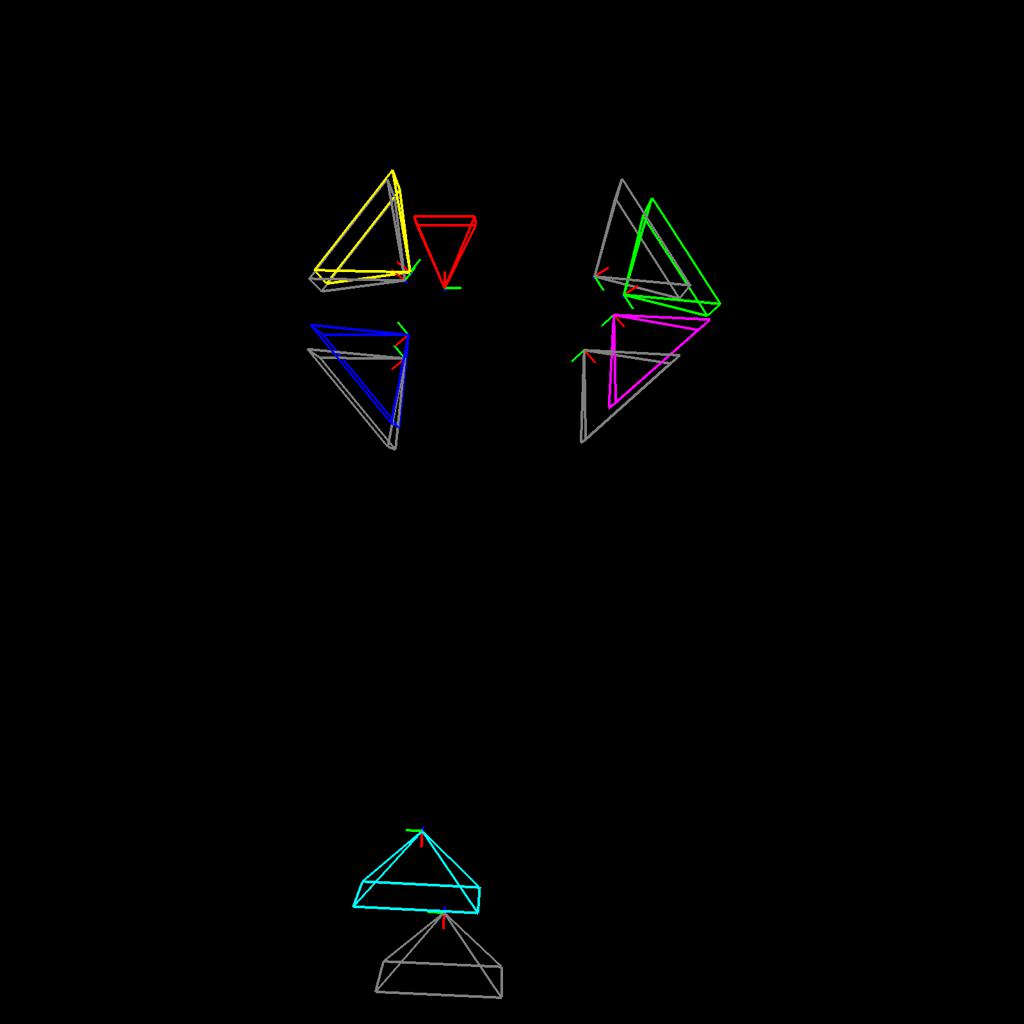}
            \includegraphics[width=0.28\textwidth]{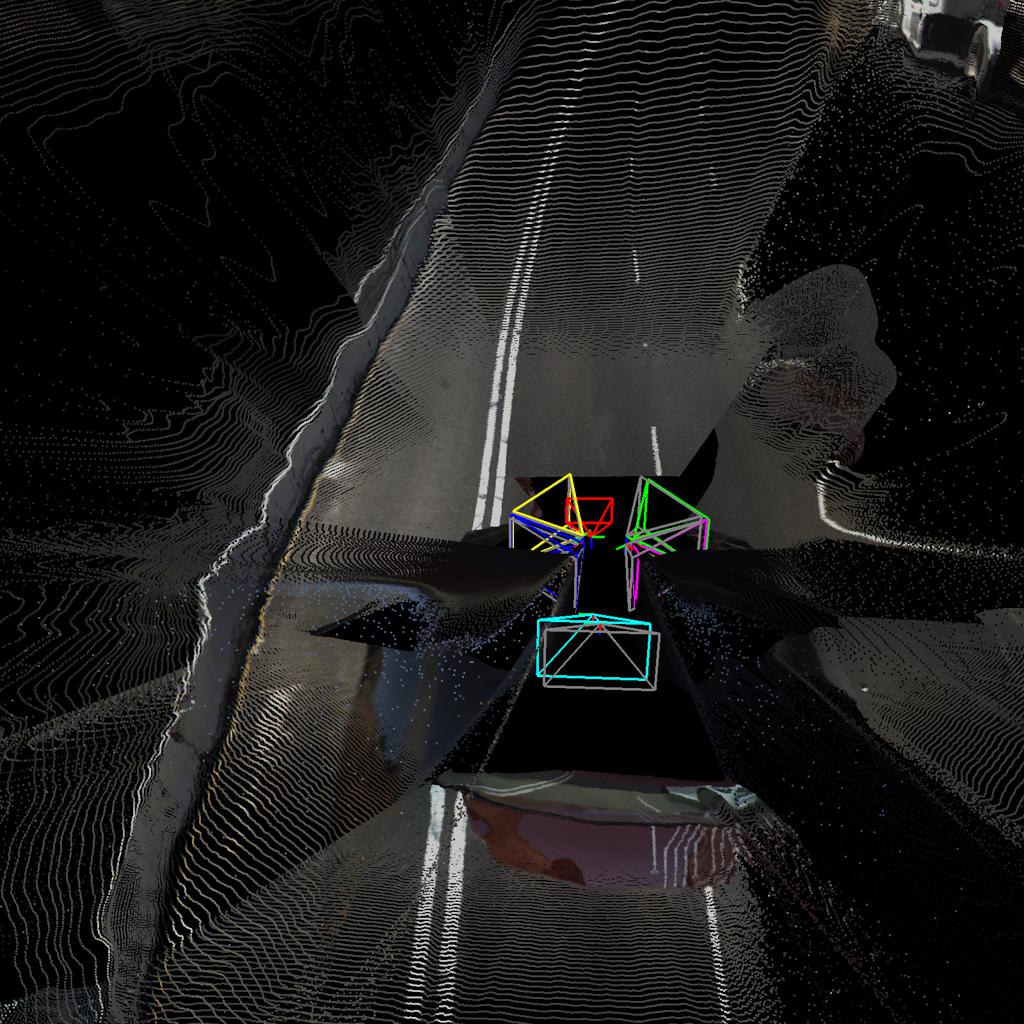}
            }
        \end{minipage}
        \hspace{-15mm}
        \begin{minipage}{.20\textwidth}
            \subfloat[Camera placement of DDAD]{
            \includegraphics[width=1.\textwidth]{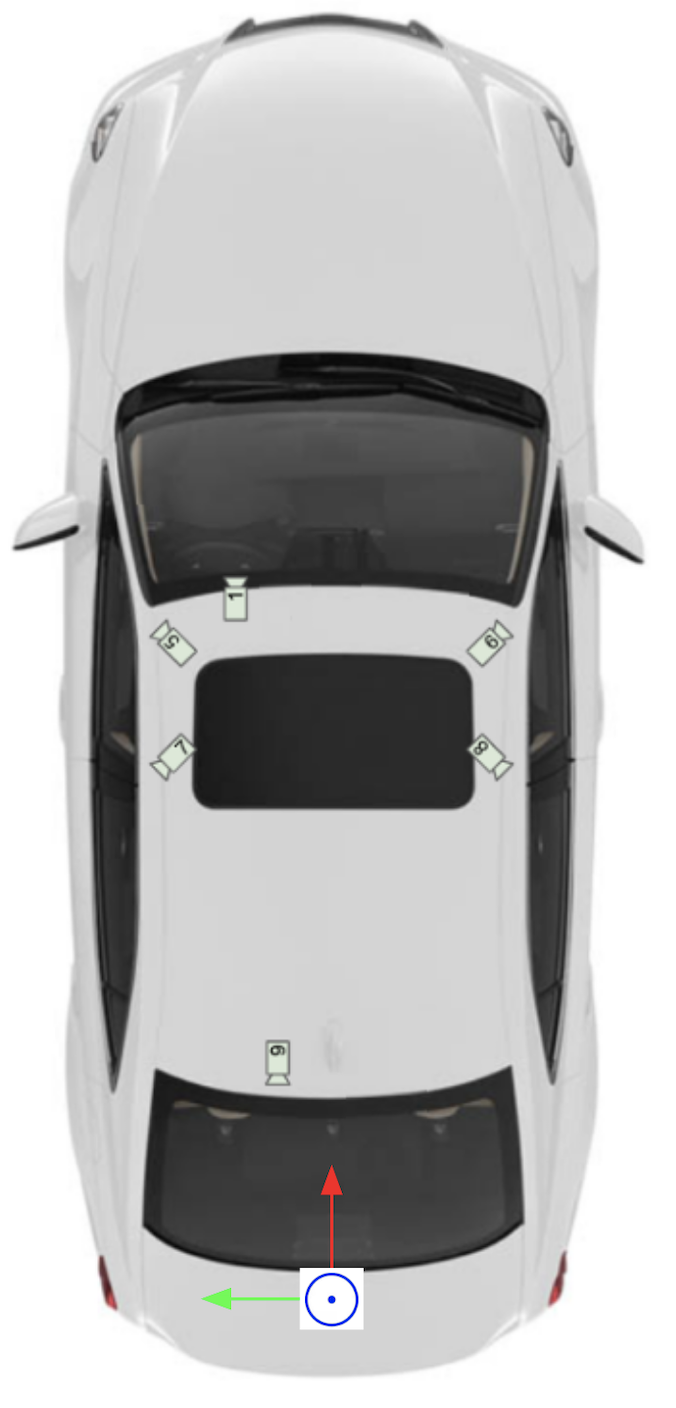}
            }
        \end{minipage}
    \end{tabular}
\caption{\textbf{Qualitative examples of calibration performance in different situations by COLMAP and \Acronym.} We show camera poses predicted by COLMAP (left) and \Acronym (center). Gray cameras are ground-truth, and colored are predictions. The projected point cloud (right) is generated using \Acronym predicted extrinsics and depth maps.}
\label{fig:ddad_calib_many}
\vspace{-3mm}
\end{figure*}

%% file: tables/vs_colmap_full.tex



\begin{table}%
\vspace{2mm}
\centering
\small
\renewcommand{\arraystretch}{1.00}
\setlength{\tabcolsep}{0.22em}
\subfloat[Translation error (Euclidean)]{
\begin{tabular}{l|ccccc|c}
    \toprule
    \multirow{2}{*}{\textbf{Method}} &
    \multicolumn{6}{c}{$t$ \lbrack m\rbrack $\downarrow$} \\
    \cmidrule{2-7}
    & \textit{F.Left}
    & \textit{F.Right}
    & \textit{B.Left}
    & \textit{B.Right}
    & \textit{Back}
    & \textit{Avg.}
    \\
    \midrule
    COLMAP
    & 
    & 
    & FAILED
    & 
    & 
    & 
    \\
    COLMAP* 
    & 6.070 
    & 2.430 
    & 2.200 
    & 2.240 
    & 2.470 
    & 3.080
    \\
    COLMAP** 
    & \underline{0.117}
    & 0.282
    & 0.210
    & 0.308
    & 0.262
    & 0.236
    \\
    \textbf{\Acronym} (\emph{I}) 
    & 0.155
    & \underline{0.240}
    & \textbf{0.167}
    & \underline{0.263}
    & \underline{0.217}
    & \underline{0.208}
    \\
    \midrule
    \midrule
    \textbf{\Acronym} 
    & \textbf{0.116}
    & \textbf{0.129}
    & \underline{0.174}
    & \textbf{0.192}
    & \textbf{0.187}
    & \textbf{0.160}
    \\
    \bottomrule
\end{tabular}
}
\\
\vspace{-2mm}
\subfloat[Rotation error (Geodesic)]{
\begin{tabular}{l|ccccc|c}
    \toprule
    \multirow{2}{*}{\textbf{Method}} &
    \multicolumn{6}{c}{$R$ \lbrack \textdegree \rbrack $\downarrow$} \\
    \cmidrule{2-7}    
    & \textit{F.Left}
    & \textit{F.Right}
    & \textit{B.Left}
    & \textit{B.Right}
    & \textit{Back}
    & \textit{Avg.}
    \\
    \midrule
    COLMAP 
    & 
    & 
    & FAILED
    & 
    & 
    & 
    \\
    COLMAP\textsuperscript{*} 
    & 2.011 
    & 1.612 
    & 1.727 
    & 1.750 
    & 1.196 
    & 1.640
    \\
    COLMAP\textsuperscript{**} 
    & \underline{1.251} 
    & \underline{0.873} 
    & \underline{1.016} 
    & \underline{0.937} 
    & \textbf{0.434} 
    & \underline{0.903}
    \\
    \textbf{\Acronym} (\emph{I}) 
    & 1.724 
    & 2.022 
    & 1.078 
    & 2.577 
    & 1.449 
    & 1.770
    \\
    \midrule
    \midrule
    \textbf{\Acronym} 
    & \textbf{0.573} 
    & \textbf{0.514} 
    & \textbf{0.576} 
    & \textbf{0.787}
    & \underline{0.628}
    & \textbf{0.616}
    \\
    \bottomrule
\end{tabular}
}
\caption{\textbf{Extrinsic calibration performance on the DDAD train split}. Since COLMAP did not converge for all sequences, we show results for sequences in which bundle adjustment (\emph{rig\_bundle\_adjuster}) converged (COLMAP\textsuperscript{*}), and in which bundle adjustment converged with translation error $t < 1$m (COLMAP\textsuperscript{**}). \AcronymInt indicates our method with additional intrinsics learning. 
}
\label{tab:full_colmap}
\vspace{-3mm}
\end{table}

%% file: tables/table_metric_ddad2.tex
\captionsetup[table]{skip=6pt}

\begin{table}[t!]
\vspace{2mm}
\renewcommand{\arraystretch}{1.00}
\centering
{
\small
\setlength{\tabcolsep}{0.2em}
\begin{tabular}{l|cccccc|c}
    \toprule
    \multirow{2}{*}{\textbf{Model}} &
    \multicolumn{7}{c}{Abs.Rel.$\downarrow$ (Metric)} \\
    \cmidrule{2-8}
    & \textit{Front}
    & \textit{F.Left}
    & \textit{F.Right}
    & \textit{B.Left}
    & \textit{B.Right}
    & \textit{Back}
    & \textit{Avg.}
    \\
    \midrule
    Monodepth+v 
    & 0.169
    & 0.203
    & 0.228
    & 0.235
    & 0.238
    & \underline{0.206}
    & 0.213 
    \\
        \textbf{\Acronym}$\dagger$ 
    & 0.165
    & \underline{0.196}
    & \underline{0.219}
    & \underline{0.216}
    & \underline{0.222}
    & 0.209
    & 0.205
    \\
    \textbf{\Acronym} 
    & \underline{0.161}
    & \textbf{0.195}
    & \textbf{0.216}
    & \textbf{0.215}
    & \textbf{0.221}
    & 0.207
    & \underline{0.202}
    \\

    \midrule
    \midrule
    FSM~\cite{fsm20222}
    & \textbf{0.130}
    & 0.201
    & 0.224
    & 0.229
    & 0.240
    & \textbf{0.186}
    & \textbf{0.201}
    \\
    \bottomrule
\end{tabular}
}
\caption{
\textbf{Metric depth estimation results on DDAD.} 
Monodepth+v is the baseline model (based on monodepth2~\cite{monodepth2}) generated by the \emph{Monodepth Pretraining}. \Acronym is the proposed method, optimized following the curriculum training procedure. \Acronym$\dagger$ is an ablation model of \Acronym, with predicted extrinsics removed. 
Note that FSM uses ground truth extrinsics for training, while our model learns them without pose supervision.}
\label{tab:depth_ddad}
\vspace{-3mm}
\end{table}

%% file: figures/stop_vehicle_000032.tex
\begin{figure}[t!] 
\centering
\vspace{2mm}
\includegraphics[width=0.32\textwidth]{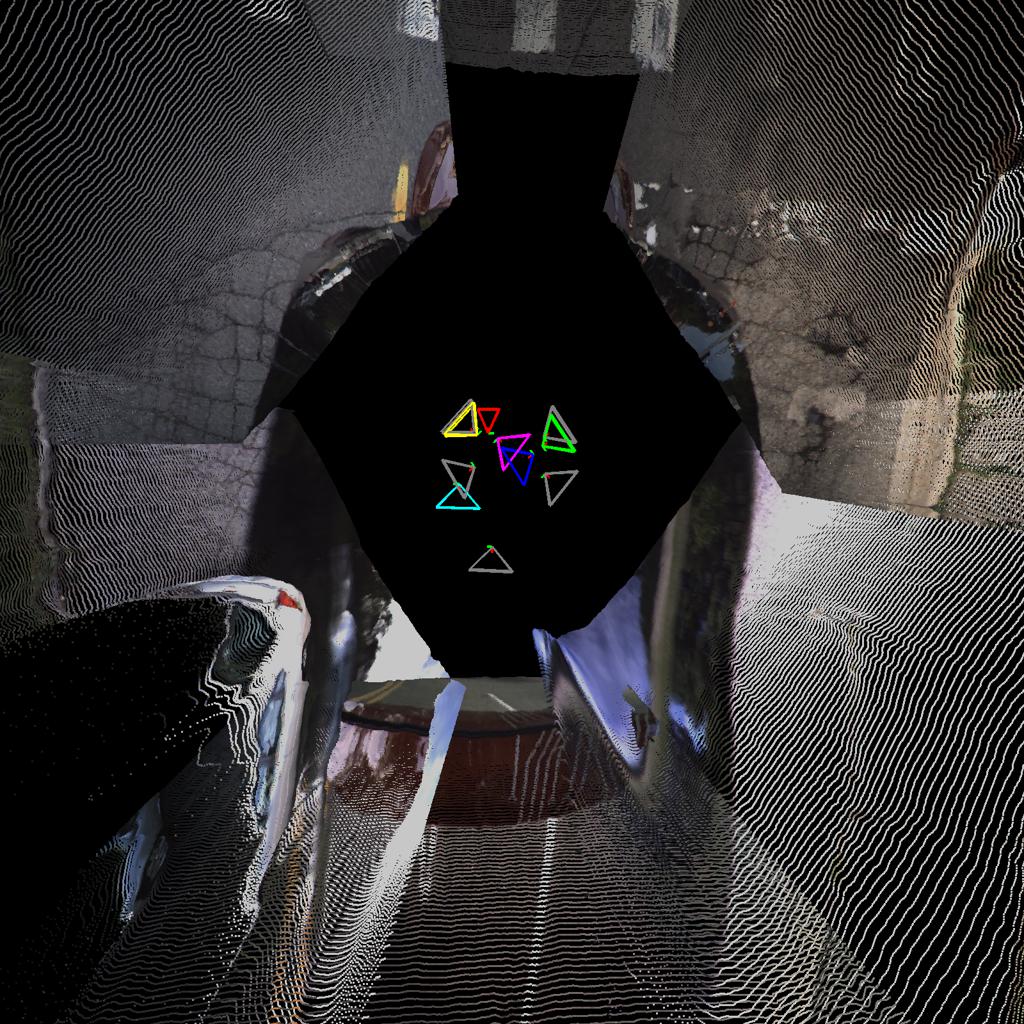}
\caption{\textbf{Example of failure case of extrinsics learning on DDAD (\emph{seq:000032}). } Three cameras (\textit{B.Left}, \textit{B.Right}, 
and \textit{Back}) failed to converge to the ground-truth position. 
In this sequence, the vehicle is stopped in front of a traffic light. 
}
\label{fig:seq52}
\vspace{-3mm}
\end{figure}

%% file: tables/curriculum_learning_ddad.tex
\begin{table}%
\vspace{2mm}
\centering
\small
\renewcommand{\arraystretch}{1.00}
\setlength{\tabcolsep}{0.22em}
\subfloat[Translation error (Euclidean)]{
\begin{tabular}{l|ccccc|c}
    \toprule
    \multirow{2}{*}{\textbf{Stage}} &
    \multicolumn{6}{c}{$t\ [m]\downarrow$} \\
    \cmidrule{2-7}
    & \textit{F.Left}
    & \textit{F.Right}
    & \textit{B.Left}
    & \textit{B.Right}
    & \textit{Back}
    & \textit{Avg.}
    \\
    \midrule 
    1. \emph{Monodepth Pre.} 
    & 0.197
    & 0.706
    & 0.476
    & 0.836
    & 1.466
    & 0.736
    \\
    2. \emph{Rotation Est.} 
    & 0.221
    & 0.678
    & 0.355
    & 0.737
    & 1.215
    & 0.641
    \\
    3. \emph{Extrinsic Est.} 
    & \underline{0.119}
    & \underline{0.130}
    & \underline{0.176}
    & \underline{0.200}
    & \underline{0.191}
    & \underline{0.163} 
    \\
    4. \emph{E2E} (\textbf{\Acronym}) 
    & \textbf{0.116}
    & \textbf{0.129}
    & \textbf{0.174}
    & \textbf{0.192}
    & \textbf{0.187}
    & \textbf{0.160}
    \\
    \midrule
    \midrule
    \emph{Extrinsic Est.} $-$ 2
    & 0.248
    & 0.478
    & 0.534 
    & 0.683 
    & 0.729 
    & 0.534 
    \\
    \emph{E2E} $-$ 3
    & 0.154
    & 0.577
    & 0.266 
    & 0.691 
    & 1.036 
    & 0.545
    \\
    \bottomrule
\end{tabular}
}
\\
\vspace{-2mm}
\subfloat[Rotation error (Geodesic)]{
\begin{tabular}{l|ccccc|c}
    \toprule
    \multirow{2}{*}{\textbf{Stage}} &
    \multicolumn{6}{c}{$R\ [^{\circ}]\downarrow$} \\
    \cmidrule{2-7}
    & \textit{F.Left}
    & \textit{F.Right}
    & \textit{B.Left}
    & \textit{B.Right}
    & \textit{Back}
    & \textit{Avg.}
    \\
    \midrule
    1. \emph{Monodepth Pre.} 
    & 50.21 
    & 56.17 
    & 121.6 
    & 127.8 
    & 2.753 
    & 71.71
    \\
    2. \emph{Rotation Est.} 
    & 1.856 
    & 1.050 
    & 2.264 
    & 2.935 
    & 1.393 
    & 1.900
    \\
    3. \emph{Extrinsic Est.} 
    & \underline{0.658}
    & \underline{0.692}
    & \underline{0.781} 
    & \underline{0.977} 
    & \underline{0.808}
    & \underline{0.783}
    \\
    4. \emph{E2E} (\textbf{\Acronym}) 
    & \textbf{0.573} 
    & \textbf{0.514}
    & \textbf{0.576} 
    & \textbf{0.787} 
    & \textbf{0.628} 
    & \textbf{0.616} 
    \\
    \midrule
    \midrule
    \emph{Extrinsic Est.} $-$ 2
    & 2.734
    & 3.969
    & 9.188 
    & 9.814 
    & 27.64
    & 10.67
    \\
    \emph{E2E} $-$ 3
    & 0.711
    & 0.938
    & 0.969 
    & 1.293 
    & 1.205 
    & 1.023
    \\
    \bottomrule
\end{tabular}
}
\caption{\textbf{Extrinsic calibration using \Acronym on DDAD}. After the \emph{Monodepth Pretraining} stage all translation and rotation vectors are initialized to zero (except the back looking camera). Next, rotation parameters are optimized in the \emph{Rotation Estimation} stage, followed by the \emph{Extrinsic Estimation} stage. Finally, a joint optimization stage (\emph{End-to-end Training}, denoted by \emph{E2E}) completes the training procedure. Skipping the \emph{Rotation Estimation} stage (denoted by \emph{Extrinsic Est.} $-$ 2) and \emph{Extrinsic Estimation} stage (denoted by \emph{E2E} $-$ 3) are shown as additional ablations. 
}

\label{tab:curriculum_l}
\end{table}


%% file: tables/calibraiton_stereo.tex
\captionsetup[table]{skip=6pt}

\begin{table}[t!]
\renewcommand{\arraystretch}{1.00}
\centering
{
\small
\setlength{\tabcolsep}{1.5em}
\begin{tabular}{l|cc}
\toprule
\textbf{Stage}
& $t$ \lbrack m\rbrack $\downarrow$
& $R$ \lbrack \textdegree \rbrack $\downarrow$
\\
\toprule
\emph{Monodepth Pretraining} 
& 0.541 
& \textbf{0.000}
\\
\emph{Rotation Estimation} 
& 0.538
& 0.201
\\
\emph{Extrinsic Estimation} 
& \underline{0.031} 
& 0.068 
\\
\emph{End-to-end Training} (\textbf{\Acronym}) 
& \textbf{0.018} 
& \underline{0.039} 

\\
\bottomrule
\end{tabular}
}
\caption{
\textbf{Extrinsic calibration evaluation via the proposed curriculum learning method on KITTI.}
The rotation parameter is also updated to avoid using prior knowledge for the stereo setup.
Thus the error $R$ is once increased from 0.000 and decreased through the optimization.
}

\label{tab:stereo_calib}
\vspace{-3mm}
\end{table}

%% file: tables/table_stereo.tex
\captionsetup[table]{skip=6pt}

\begin{table}[t!]
\vspace{2mm}
\renewcommand{\arraystretch}{1.00}
\centering
{
\small
\setlength{\tabcolsep}{0.3em}
\begin{tabular}{l|cccc}
\toprule
\textbf{Method} & 
Abs Rel$\downarrow$ &
Sq Rel$\downarrow$ &
RMSE$\downarrow$ &
$\delta_{1.25}$ $\uparrow$
\\
\toprule

\textbf{SESC}$\dagger$
& 0.114 
& 0.831 
& 4.788 
& \underline{0.863}
\\
\textbf{\Acronym}
& 0.113 
& \underline{0.819}
& 4.777 
& 0.864
\\
\midrule
\midrule
PackNet-SfM (M+v)~\cite{packnet}
& \underline{0.111} 
& 0.829 
& 4.788 
& 0.864 
\\
FisheyeDistanceNet~\cite{fisheye2020}
& 0.117 
& 0.867 
& \underline{4.739} 
& 0.869 
\\
FSM~\cite{fsm20222}
& \textbf{0.108} 
& \textbf{0.737} 
& \textbf{4.615} 
& \textbf{0.872}
\\
\bottomrule
\end{tabular}
}
\caption{
\textbf{Depth estimation results by scale-aware models on KITTI.} Our proposal \Acronym is competitive to the other state-of-the-art methods. \Acronym$\dagger$\ is an ablation of \Acronym, trained without the predicted extrinsics while the other training configuration is kept the same. 
Note that FSM~\cite{fsm20222} uses ground truth extrinsics (stereo information) for training. Contrary, our model predicts that extrinsics. 
}
\label{tab:stereo}
\end{table}

%% file: sections/conclusion.tex
We propose a novel methodology for camera extrinsic calibration based on self-supervised depth and ego-motion learning, focusing on efficiency and robustness. 
Our experimental evaluation on driving benchmarks demonstrates that applying our proposed curriculum learning strategy, composed of \emph{Monodepth Pretraining},  \emph{Rotation Estimation}, \emph{Extrinsic Estimation}, and \emph{End-to-end Training}, makes it possible to simultaneously learn depth, ego-motion and extrinsic parameters over more than one hundred data collection sequences, while achieving performance comparable to traditional structure-from-motion systems. Furthermore, depth estimation experiments demonstrate that our predicted extrinsics can be used to improve depth estimation performance without requiring additional information in the form of ground-truth extrinsics.